\documentclass{article}

\usepackage[utf8]{inputenc}  
\usepackage[T1]{fontenc}     
\usepackage[margin=0.75in]{geometry}
\usepackage{amsmath}
\usepackage{graphicx}

\usepackage{array} 
\usepackage{enumitem}
\usepackage{booktabs}
\usepackage{hyperref}

\usepackage{authblk}

\usepackage{listings}
\title{GAAPO: Genetic Algorithmic Applied to Prompt Optimization}
\author[1]{Xavier Sécheresse}
\author[1]{Jacques-Yves Guilbert--Ly}
\author[1]{Antoine Villedieu de Torcy}
\affil[1]{Biolevate}

\date{\today} 

\begin{document}

\maketitle



\section*{Abstract}
Large Language Models (LLMs) have demonstrated remarkable capabilities across various tasks, with their performance heavily dependent on the quality of input prompts \cite{schulhoff2025promptsurvey} \cite{sahoo2025promptengineering}. While prompt engineering has proven effective, it typically relies on manual adjustments, making it time-consuming and potentially suboptimal. This paper introduces GAAPO (Genetic Algorithm Applied to Prompt Optimization), a novel hybrid optimization framework that leverages genetic algorithm \cite{dejong1988gen} principles to evolve prompts through successive generations. Unlike traditional genetic approaches that rely solely on mutation and crossover operations, GAAPO integrates multiple specialized prompt generation strategies within its evolutionary framework. Through extensive experimentation on diverse datasets including ETHOS, MMLU-Pro, and GPQA, our analysis reveals several important point for the future development of automatic prompt optimization methods: importance of the tradeoff between the population size and the number of generations, effect of selection methods on stability results, capacity of different LLMs and especially reasoning models to be able to automatically generate prompts from similar queries... Furthermore, we provide insights into the relative effectiveness of different prompt generation strategies and their evolution across optimization phases. These findings contribute to both the theoretical understanding of prompt optimization and practical applications in improving LLM performance.

\section{Introduction} \label{section:introduction}
Large Language Models (LLMs) have gained significant attention following the public release of generative AI assistants such as ChatGPT (2022) and Claude (2023). A critical factor in maximizing these models' effectiveness lies in the quality of input prompts - the instructions that guide LLMs toward generating relevant outputs. While the impact of prompting on LLM performance has been well-documented through various benchmarks \cite{schulhoff2025promptsurvey}, the process typically relies on manual adjustments, making it both time-consuming and susceptible to human error. This highlights the necessity for developing automated methods to fully harness the capabilities of modern LLMs.

In response to this need, several machine learning approaches have been developed to automate prompt optimization. Reinforcement learning has been employed to optimize evaluation costs and computational efficiency \cite{yang2024opro} \cite{ma2024eureka}, while in-context learning focuses on improving prompt performance through example-based learning \cite{dong2024surveyICL}. Regression techniques have been explored to establish direct relationships between prompt characteristics and model performance \cite{feffer2024promptreg}. These diverse approaches aim to streamline the prompting process, reducing the reliance on manual intervention while addressing different aspects of prompt optimization.

Recent research has shown that smaller language models can achieve performance comparable to larger LLMs through various optimization techniques such as distillation \cite{xu2024surveydistil} and prompt engineering \cite{schulhoff2025promptsurvey}. While traditional approaches like distillation modify model weights, prompt optimization offers a more flexible alternative: it enhances model performance without altering the underlying architecture. This approach is particularly valuable as it can be applied to any LLM regardless of size or architecture, providing a generalizable framework for task-specific optimization while maintaining cost-effectiveness.

In this work,  we introduce GAAPO (Genetic Algorithmic Applied to Prompt Optimization), an algorithm that integrates different prompt generation strategies into a hybrid prompt optimizer. This innovative approach capitalizes on the strengths of diverse techniques, ensuring optimal performance. Crucially, it maintains a detailed record of the evolution of prompting strategies, which is essential for tracking progress and making informed adjustments. The design of this optimizer prioritizes adaptability, ensuring it can seamlessly incorporate future advancements in the field, thereby remaining relevant and effective as new techniques and models emerge.
\section{Related works} \label{section:related_works}
\subsection{Prompt Engineering}

Prompt engineering is a critical aspect of working with large language models (LLMs), as it involves crafting inputs that guide the model to produce desired outputs. It has been demonstrated that this step is critical to enhance LLM capabilities \cite{schulhoff2025promptsurvey}. However, this process requires a deep understanding of both the model's capabilities and the specific task at hand. Traditionally, prompt engineering has been a manual process \cite{bsharat2024handprompt}, relying on human intuition and expertise to iteratively refine prompts for optimal performance.

\subsection{Automatic Prompt Engineering}

The limitations of manual prompt engineering have led to the development of automated approaches. These methods utilize machine learning algorithms to automatically generate and optimize prompts, reducing the need for manual intervention and democratizing access to advanced language processing capabilities. To standardize these developments, frameworks like DSPy \cite{khattab2024dspy} have emerged, providing a systematic approach to develop and evaluate automatic prompt optimization methods.
Various approaches have been explored in this field, from "gradient-oriented" prompt evolution \cite{shin2020autoprompt} to more sophisticated optimization techniques. Notable advances include APO \cite{pryzant2023apo}, which introduced gradient-based prompt optimization, while OPRO \cite{yang2024opro} demonstrated the effectiveness of using LLMs themselves as optimizers. These automated methods can efficiently explore vast prompt spaces, identifying optimal prompts that maximize model performance on specific tasks. This systematic approach has become increasingly important as LLMs are deployed in diverse applications, where task-specific prompt optimization can significantly impact performance.

\begin{figure}[h]
    \centering
    \fbox{\includegraphics[width=0.8\textwidth]{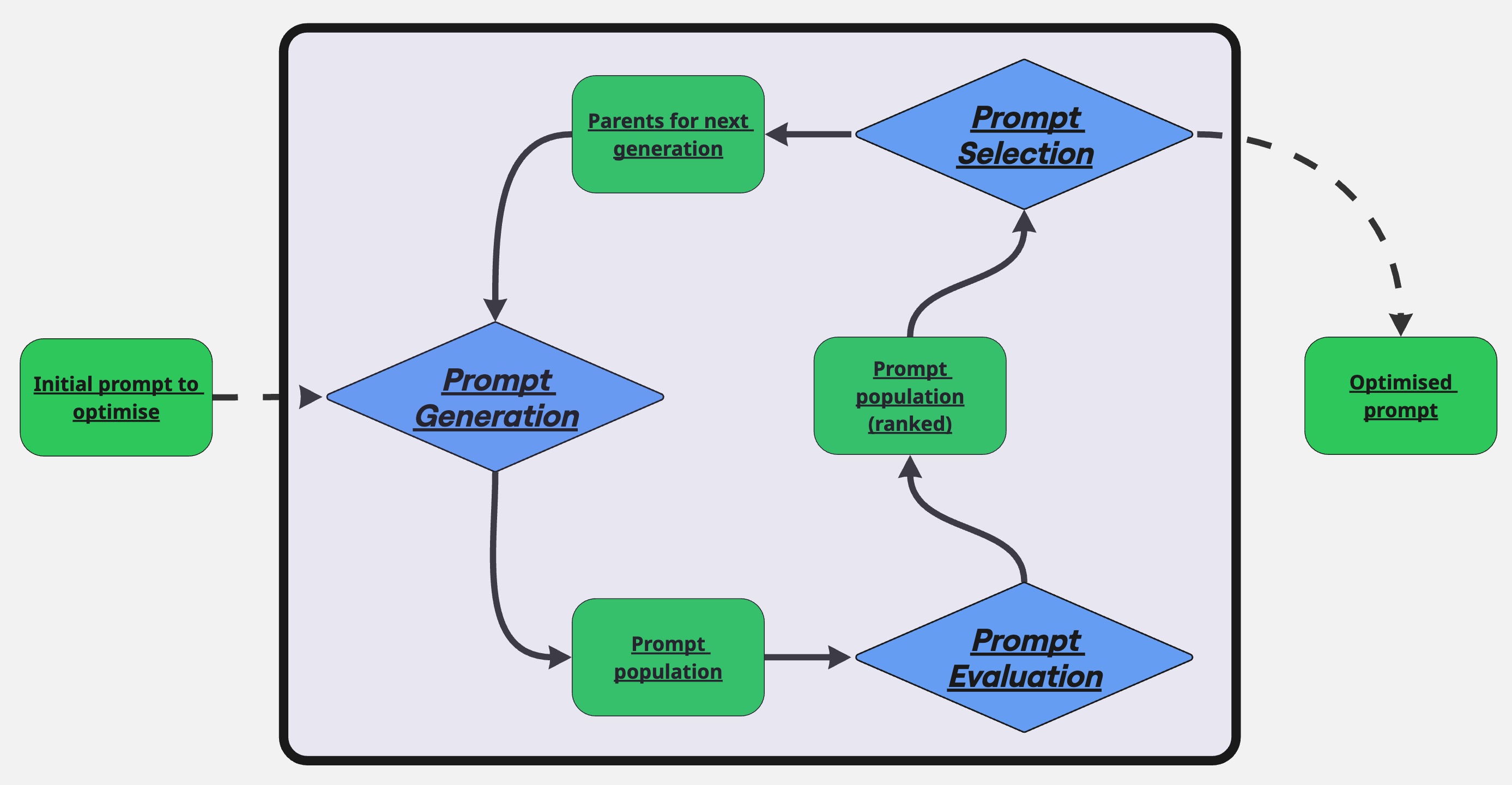}}
    \caption{Schema of the general automatic prompt optimization process}
    \label{fig:genmodels}
\end{figure}

Most prompt optimization techniques follow the same architecture, described in the figure \ref{fig:genmodels}.

\subsection{Genetic algorithm}

Genetic algorithms (GAs) are a class of optimization techniques inspired by the principles of natural selection and genetics \cite{dejong1988gen}. By mimicking the evolutionary process, GAs have been successfully applied to various machine learning and artificial intelligence tasks \cite{lambora2019genalgo}. They are particularly effective in solving complex optimization problems where traditional methods struggle, thanks to their ability to explore large and poorly understood search spaces. The GA process begins with a randomly initialized population of candidate solutions, each evaluated based on a fitness function that measures its effectiveness in solving the problem. The best-performing individuals are selected for reproduction using evolutionary operators such as crossover, which recombines elements from two solutions, and mutation, which introduces random modifications to enhance diversity.

Over the years, GAs have been widely adopted in nearly every field of machine learning, including feature selection \cite{yang1998feature}, neural network optimization \cite{stanley2002evolving}, hyperparameter tuning \cite{beielstein2003evolutionary}, clustering \cite{maulik2000genetic}, and reinforcement learning \cite{whiteson2006evolutionary}. For example, GAs have been used to optimize neural network architectures by evolving network topologies and weight configurations, improving model performance and efficiency \cite{stanley2002evolving}. In reinforcement learning, they have been leveraged to evolve policies and reward functions, enabling agents to learn complex behaviors \cite{whiteson2006evolutionary}. Additionally, hybrid approaches combining GAs with local search techniques have been developed to improve convergence speed and accuracy \cite{merz2000genetic}. Parallel implementations of GAs further enhance their scalability, allowing them to tackle large-scale optimization problems efficiently \cite{cantupaz2001efficient}. The adaptability and robustness of genetic algorithms make them a powerful tool for advancing machine learning methodologies and solving a diverse range of computational challenges.

\subsection{Application to prompt optimization}

Genetic algorithms have been previously explored in prompt optimization, though their implementations often focus on specific aspects of the prompt space. EvoPrompt \cite{guo2024evoprompt} introduces a basic evolutionary approach where new prompts are primarily generated through crossover operations, combining successful segments from parent prompts followed by linguistic refinement. This method, while effective, primarily explores structural variations within a limited scope of the prompt space.
A more sophisticated approach is demonstrated by PhaseEvo \cite{cui2024phasevo}, which implements a two-phase evolutionary strategy. The first phase employs global mutations to identify promising regions in the prompt space, effectively searching for potential global optima. The second phase then applies more focused optimizations through semantic mutations and gradient-based refinements. 

However, despite their innovative contributions, these approaches operate within relatively narrow paradigms of prompt generation and modification. While they effectively handle structural and semantic modifications, they don't fully explore the broader spectrum of prompt transformation strategies. Moreover, they lack a comprehensive framework that could integrate existing prompt optimization techniques or adapt to emerging methodologies in the field. This limitation in extensibility and modularity restricts their ability to evolve alongside new developments in prompt engineering.

\section{Data} \label{section:data}

\underline{\textbf{ETHOS dataset}}

The ETHOS (Ethics in Text - Hate and Offensive Speech) multilabel dataset \cite{mollas2022ethos} is a specialized benchmark designed to evaluate hate speech recognition capabilities in language models. It consists of 443 carefully annotated text samples categorized across eight distinct dimensions of hate speech and offensive content, including race, gender, and violence. Each sample in the dataset is labeled to indicate the presence or absence of specific types of harmful content, enabling fine-grained evaluation of model performance in detecting various forms of hate speech. The dataset's multi-label structure allows for comprehensive assessment of language models' ability to identify intersecting forms of discriminatory or offensive content, making it particularly valuable for evaluating ethical content moderation capabilities. The balanced distribution across different categories of hate speech ensures robust evaluation across the spectrum of harmful content typically encountered in real-world applications. \\

\noindent\underline{\textbf{Complementary datasets}}

To assess performances of our approach on a wide range of tasks, we evaluated our model on 3 other datasets, alongside with ETHOS-multilabel:
\begin{itemize}
    \item The MMLU-Pro (Massive Multitask Language Understanding Professional)\cite{wang2024mmlu} dataset extends the famous MMLU \cite{hendrycks2021mmlu} dataset by complexifying it to a professional level, with our focus on two key subcategories. The Engineering subcategory evaluates technical understanding across various engineering disciplines, testing knowledge of fundamental principles, technical specifications, and complex problem-solving approaches encountered in professional practice. The Business subcategory assesses comprehension of management principles, corporate strategy, financial decision-making, and organizational behavior through practical business scenarios. Repartition of MMLU-Pro dataset in subcategories is detailed in annex \ref{annex:mmlupro}.
    \item GPQA (General Physics Question Answering)\cite{rein2023gpqa} presents a specialized evaluation framework for physics understanding through multiple-choice questions. The dataset covers a broad spectrum of physics topics, from mechanics to quantum physics, requiring both theoretical knowledge and practical problem-solving abilities. Questions are designed to test not only recall of physical principles but also their application in solving concrete problems, making it an effective benchmark for assessing scientific reasoning capabilities in LLMs.
\end{itemize}

\section{Methods} \label{section:methods}
\subsection{Models}

\subsubsection{GAAPO: Genetic Algorithmic Applied to Prompt Optimization}

\begin{figure}[h]
    \centering
    
    \includegraphics[width=0.98\textwidth]{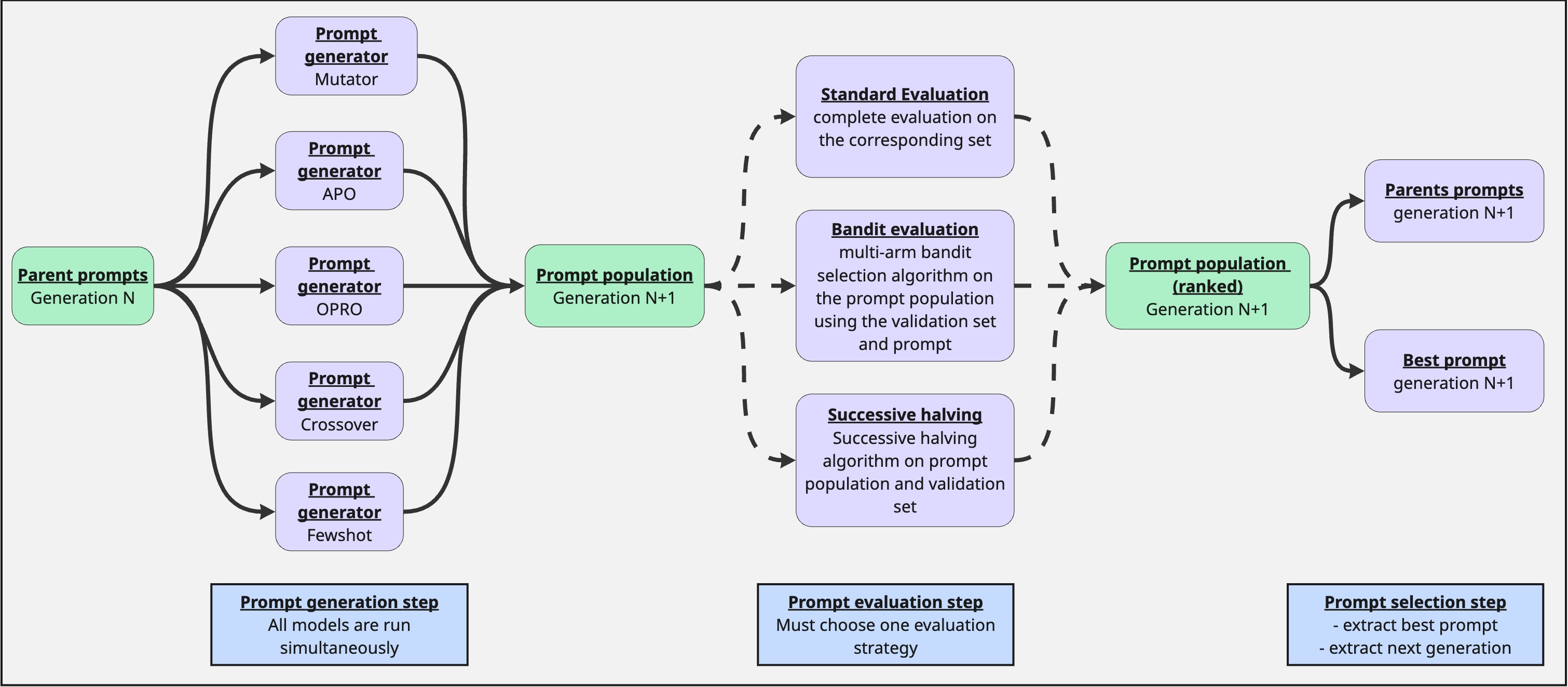}
    \caption{Description of the GAAPO optimization process.}
    \label{fig:gaapo}
\end{figure}

GAAPO (Genetic Algorithm Applied to Prompt Optimization) follows the principles of genetic algorithms to evolve and optimize prompts through successive generations. The algorithm combines multiple prompt optimization strategies to explore a broader prompt space than previous methods, leveraging the strengths of each approach while maintaining the evolutionary nature of genetic algorithms.
The optimization pipeline, inspired by existing works \cite{pryzant2023apo}\cite{cui2025mapo} and described in the figure \ref{fig:gaapo}, operates in three distinct phases during each generation:

\begin{itemize}[noitemsep, topsep=0pt]
    \item Generation phase: New prompt candidates are created using multiple strategies, with each strategy operating on a subset of high-performing prompts from the previous generations.
    \item Evaluation phase: The newly generated population is evaluated on the validation set using either exhaustive evaluation or a bandit-based approach to optimize computational resources.
    \item Selection phase: Top-performing prompts are selected based on their evaluation scores to serve as parents for the next generation, ensuring best performers are used as parents at all time for the future generations.
\end{itemize}

This iterative process combines the exploration capabilities of genetic algorithms with specialized prompt optimization techniques, enabling efficient navigation of the prompt space while maintaining diversity in the population. \\

\noindent\textbf{\underline{Prompt generation}}

The genetic algorithm framework incorporates multiple prompt generation methods, each implementing distinct optimization strategies as detailed in sections \ref{subsection:prompt_gen2} and \ref{subsection:prompt_gen1}. These methods, summarized in Table \ref{tab:prompt_generation}, represent diverse approaches to prompt optimization, each with its own strengths and limitations. The hybrid nature of GAAPO leverages this diversity by combining these complementary strategies within a single optimization framework. This integration enables the algorithm to capitalize on the advantages of each method while mitigating their individual limitations through iterative application of varied optimization approaches. The synergistic combination of these methods allows for more comprehensive exploration of the prompt space than would be possible with any single strategy. Examples of generated prompts on a real-world task are presented in annex \ref{annex:prompt_obtained}.

To streamline the optimization process, we unified the selection and evaluation phases across all different optimization methods into a single coherent framework. This architectural decision maintains only the generative (expansion) phases of these algorithms, integrating them as candidate generation strategies within GAAPO's evolutionary cycle. This simplification allows for consistent evaluation metrics and selection criteria across all generated candidates while preserving the unique prompt generation characteristics of each method.

Compared to already existing GA-related prompt optimization methods, this framework allows a wider exploration of the prompt space, leveraging advantages of all implemented methods and not focusing on single-algorithm local improvements. \\

\noindent\textbf{\underline{Evaluation}}

To meaningfully compare new prompts, we evaluate them on a subset of the task we have at hand and compare their accuracy (in the current setting).

Several strategies has been implemented for the evaluation process to rank the individuals in each generation:
\begin{itemize}[noitemsep, topsep=0pt]
    \item Complete evaluation: Run a standard evaluation of each prompt on the evaluation set and rank new prompts according to their accuracy.
    \item Successive halving (SH) process \cite{schmucker2021sh}:  prompt accuracies are compared on a subset of the dataset, the top-performing half of the models is retained, and the survivors are evaluated on a new subset. This process is repeated iteratively until only a few models remain. This approach allows to drastically reduce the number of API calls but increases the risk to remove interesting prompts from the evaluation very early due to the disparity of evaluations results on subsets.
    \item Bandit selection algorithm\cite{slivkins2024bandit}: run a multi-arm selection bandit algorithm. Evaluate subsets of the prompt population on batches of data, and apply the UCB-E reward model \cite{han2024ucbbandits} to identify the best arms. Note that this method was also used in the original paper of APO \cite{pryzant2023apo}.
\end{itemize}
\strut

\noindent\textbf{\underline{Selection}}

The selection step used to generate the new parents at each generation is quite simple. We simply chose, among all the prompts which have been evaluated, the best according to their evaluation score.

\subsubsection{Generation methods: "forced" evolutions} \label{subsection:prompt_gen2}

The first categories of generators were directly inspired from standard prompt optimization models, described below. This methods directly use previous prompts to generate new ones, by using the errors made (APO) or trying to expand a prompt trajectory (OPRO), hence the "forced" evolution. \\

\noindent\textbf{\underline{OPRO:}} Optimization by PROmpting
\vspace{5pt}

OPRO (Optimization by PROmpting) \cite{yang2024opro} is an iterative prompt optimization algorithm that leverages large language models to generate and refine prompts through a trajectory-based optimization approach. The algorithm maintains a trajectory of the top-performing prompts, ranked by their performance scores, and uses this historical information to guide the generation of new candidates.
During training, OPRO employs a stochastic dropout mechanism on the trajectory of best-performing prompts to maintain diversity and prevent convergence to local optima. The filtered trajectory then serves as input for the generation of new candidate prompts, which are subsequently evaluated on the current set. This evaluation process updates the trajectory, maintaining a dynamic optimization path. \\

\noindent\textbf{\underline{ProTeGi:}} Prompt Optimization with Textual
Gradients
\vspace{5pt}

\begin{figure}[h]
    \centering
    \includegraphics[width=0.98\textwidth]{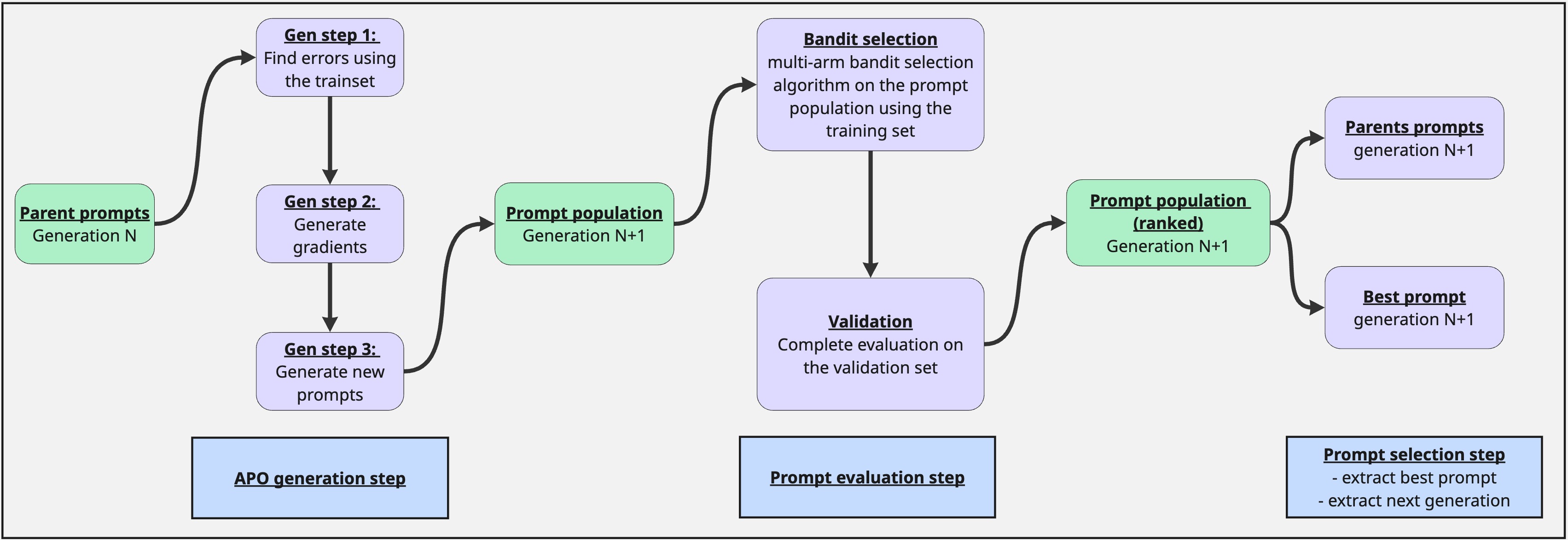}
    \caption{Description of the APO optimisation process, which served as a basis for GAAPO.}
    \label{fig:apo}
\end{figure}

The Automatic Prompt Optimizer (APO/ProTeGi) \cite{pryzant2023apo} is an iterative algorithm designed to automatically optimize prompts for Large Language Models through a three-phase process described in figure \ref{fig:apo}. The expansion phase begins by evaluating existing prompts to identify errors, which are then grouped for focused analysis. The algorithm generates improvement "gradients" from these errors and creates new candidate prompts. In the selection phase, APO employs multi-armed bandit strategies (such as epsilon-greedy \cite{kuleshov2014banditreview} or Bayesian UCB \cite{han2024ucbbandits}) to efficiently identify promising candidates. This approach balances the exploration of new prompt variations with the exploitation of proven patterns, evaluating candidates on small batches for computational efficiency. The validation phase assesses selected candidates on a separate validation set ensuring the robustness of the optimized prompts. Key features include parallel processing, adaptive error analysis, and gradient-guided refinement.

\subsubsection{Generation methods: random evolutions} \label{subsection:prompt_gen1}

To complement the "forced" evolution optimization strategies, we developed three additional prompt generation methods that were incorporated into GAAPO's framework. These supplementary approaches expand the algorithm's capacity to explore diverse regions of the prompt space. They all use already existing prompts to generate new ones by randomly modifying them.\\

\noindent\textbf{\underline{Random Mutator}}: Prompt random mutation
\vspace{5pt}

The Random Mutator serves as a mutation operator within a genetic algorithm framework, designed to explore the vast prompt space through controlled random modifications. This approach draws inspiration from biological mutations in genetic evolution, where random changes can lead to beneficial adaptations. The mutation process operates by randomly selecting from eight distinct mutation strategies, each targeting different aspects of prompt engineering: 
\begin{itemize}[noitemsep, topsep=0pt]
    \item \textbf{instruction expansion}: adds detailed guidelines,
    \item \textbf{expert persona injection}: introduces specialized viewpoints,
    \item \textbf{structural variation}: modifies the prompt's architecture,
    \item \textbf{constraint addition}: introduces new boundaries,
    \item \textbf{creative backstory}: weaves narrative elements,
    \item \textbf{task decomposition}: breaks down complex instructions,
    \item \textbf{concise optimization}: streamlines the content, 
    \item \textbf{role assignment}:  establishes specific model behaviors.
\end{itemize}

Each mutation creates a new variant of the original prompt (examples are presented in annex \ref{annex:prompt}), potentially discovering more effective formulations. Like genetic mutations in nature, these modifications can range from subtle adjustments to significant transformations, allowing for both local and global exploration of the prompt space. This random but structured approach enables the discovery of novel prompt variations that might not be obvious through deterministic methods.\\

\noindent\textbf{\underline{Crossover:}} random prompt merging
\vspace{5pt}

Crossover operations in prompt engineering also draw inspiration from genetic algorithms' recombination mechanisms, but require careful adaptation for text-based prompts. While traditional genetic algorithms can perform straightforward splitting and merging of genetic sequences, prompt crossover needs to maintain semantic coherence and structural integrity. In our implementation, we developed a simple yet effective crossover mechanism: given two parent prompts that have demonstrated good performance, the operation splits each prompt approximately at its midpoint and combines the first half of one prompt with the second half of the other. This approach, while basic (and which could be optimized), provides several advantages:
\begin{itemize}[noitemsep, topsep=0pt]
    \item It preserves coherent instruction blocks from each parent
    \item It enables the combination of different strategic elements (e.g., merging a prompt with strong reasoning guidelines with another that has effective constraint definitions)
    \item It maintains a balance between exploration and preservation of successful prompt components
\end{itemize}

However, this straightforward approach could be enhanced in future work through more sophisticated crossover mechanisms, such as semantic block identification and recombination, or intelligent selection of crossover points based on prompt structure analysis.

\noindent\textbf{Note:} Already existing GA-related prompt optimization methods such as EvoPrompt \cite{guo2024evoprompt} are a combination of these first two categories of methods.\\

\noindent\underline{\textbf{Fewshot:}} In-context learning for prompt optimization
\vspace{5pt}

In-context learning is a fundamental capability of large language models (LLMs)\cite{dong2024surveyICL} that allows them to adapt their behavior based on examples provided within the prompt, without requiring model parameter updates. This ability enables LLMs to understand and emulate patterns from demonstrated examples in real-time.
The few-shot algorithm for prompt optimization leverages this capability by augmenting existing prompts with selected examples while maintaining the original prompt's structure and purpose. The process begins by randomly selecting 1 to 3 labeled examples from the training dataset for each parent prompt. These examples are then appended to the original prompt in a structured format, with clear input-output pairs. The algorithm is computationally efficient as it doesn't require complex prompt modifications or extensive evaluations. Instead, it relies on the natural ability of LLMs to learn from examples, making it a practical approach for prompt enhancement while maintaining the original prompt's core functionality.

\begin{table}[h]
\centering
\begin{tabular}{|p{2cm}|p{6cm}|p{6cm}|}
\hline
\textbf{Method} & \textbf{Advantages} & \textbf{Drawbacks} \\
\hline
Mutations & 
\begin{itemize}[nosep]
    \item Simple and efficient implementation
    \item Multiple mutation strategies available
    \item Maintains prompt diversity
    \item Low computational cost
\end{itemize} & 
\begin{itemize}[nosep]
    \item Can produce invalid prompts
    \item Changes might be too random
    \item Limited by predefined mutation strategies
\end{itemize} \\
\hline
APO & 
\begin{itemize}[nosep]
    \item Error-driven optimization
    \item Targeted improvements based on failure analysis
    \item Systematic approach to prompt refinement
\end{itemize} & 
\begin{itemize}[nosep]
    \item Computationally expensive
    \item Requires error examples
    \item May overfit to specific error patterns
\end{itemize} \\
\hline
OPRO & 
\begin{itemize}[nosep]
    \item Learns from successful prompts
    \item Efficient use of historical information
\end{itemize} & 
\begin{itemize}[nosep]
    \item Dependent on quality of previous generations
    \item Can converge to local optima
    \item Higher LLM usage per generation
\end{itemize} \\
\hline
Crossover & 
\begin{itemize}[nosep]
    \item Combines successful prompt features
    \item Preserves effective components
    \item Low computational cost
\end{itemize} & 
\begin{itemize}[nosep]
    \item Simple splitting might break prompt coherence
    \item Requires multiple good parents
    \item Can produce semantically invalid combinations
\end{itemize} \\
\hline
FewShot & 
\begin{itemize}[nosep]
    \item Improves prompt with concrete examples
    \item Helps model understand edge cases
    \item Direct performance feedback
\end{itemize} & 
\begin{itemize}[nosep]
    \item Can make prompts too lengthy
    \item Risk of overfitting to examples
    \item Limited by example quality and availability
\end{itemize} \\
\hline
\end{tabular}
\caption{Comparison of Prompt Generation Methods in Hybrid Genetic Optimizer}
\label{tab:prompt_generation}
\end{table}

\subsubsection{Examples}
Table \ref{tab:obtained_prompts} presents prompts obtained by several methodologies on the ETHOS multilabel dataset after 10 generations. The original prompt from where all optimization trajetories started is: 

\begin{lstlisting}
A message from a user, your goal is to determine if this message is a hate speech or not: user_message. Possible class for the hate speech are: violence, directed_vs_generalized, gender, race, national_origin, disability, religion, sexual_orientation.
\end{lstlisting}

\begin{table}[h!]
    \centering
    \begin{tabular}{|l|p{16cm}|}
        \hline
        \textbf{Method} & \textbf{Obtained Prompt} \\
        \hline
        
        \hline 
        APO &
        \begin{lstlisting}
Analyze the given message to determine if it contains hate speech: user_message.
Definition: Hate speech is defined as language that attacks or diminishes a person or group based on attributes such as race, religion, sexual orientation, gender, disability, national origin, or other protected characteristics. Your task is to:
- Provide a detailed reasoning justifying your classification.
- Categorize the message into the most relevant categories from the following list. Each category should be clearly justified in your reasoning:
+ sexual_orientation: Attacks or offensive language targeting individuals based on their sexual orientation or gender identity.
+ religion: Language that attacks or demeans individuals based on their religious beliefs or affiliation.
+ [... definitions of other categories...]
        \end{lstlisting} \\
        \hline
        OPRO &
        \begin{lstlisting}
Analyze the message to determine if it is hate speech: user_message.
Consider the following factors in your analysis:
- Presence of derogatory language or slurs targeting race, gender, religion, or other protected characteristics.
- Any explicit or implicit threats, calls to violence, or incitement of harm.
- Discriminatory statements that demean or degrade individuals or groups.
- The overall tone, intent, and context of the message.
Ensure your reasoning is thorough and your solution is actionable and clear. Possible categories for hate speeches are: violence, directed_vs_generalized, gender, race, national_origin, disability, religion, sexual_orientation.
        \end{lstlisting}\\
        \hline
        Mutator & \begin{lstlisting} 
Analyze the message to determine if it is hate speech, using the following structured approach: 
- Identify offensive language targeting protected characteristics.
- Assess intent and context.
- Evaluate potential harm.
- Ensure bias-free analysis adhering to hate speech policies.
Provide evidence-based reasoning and actionable solutions. Examples: Racial slurs dehumanize based on race; messages promoting violence incite harm; derogatory gender terms cause marginalization. Analyze: user_message. 
         \end{lstlisting}\\
        \hline 
        \textbf{Ours} & \begin{lstlisting} 
You are a team of experts tasked with determining if a given message contains hate speech. Your team consists of three roles:
**The Critic**: Your first responsibility is to carefully analyze the message for any indicators of hate speech, such as offensive language, derogatory terms, or discriminatory content. Consider the context and tone of the message.
**The Problem Solver**: After the critic has provided their analysis, your role is to determine whether the message qualifies as hate speech based on the indicators identified. Consider the broader implications of labeling the message as hate speech.
**The Expert Reviewer**: Your role is to review the analysis and solution provided by the previous roles.
Ensure that the reasoning is thorough, the solution is accurate, and the output is consistent with the examples provided. Here is the message to analyze: user_message.
        \end{lstlisting} \\
    \hline
    \end{tabular}
    \caption{Prompts obtained via different optimization methods on the ETHOS multilabel dataset}
    \label{tab:obtained_prompts}
\end{table}

\subsection{Optimization framework}

HOPR (Hint Optimization and Prompt Refinement) is a Python framework designed for systematic prompt optimization and evaluation. Like DSPy \cite{khattab2024dspy}, it provides a structured approach to prompt engineering, but with a distinct focus on evolutionary optimization techniques. While DSPy emphasizes the composition and chaining of language model operations through programmatic interfaces, HOPR specializes in automated prompt optimization through a variety of implemented strategies extracted from the state of the art methods for automatic prompt engineering.

The framework is built around modular components: optimizers that implement different prompt generation strategies, metrics for evaluation, and a core system for managing prompt evolution. HOPR's architecture allows researchers to easily implement and compare different prompt optimization techniques, track the evolution of prompts to study the best optimization methods, and maintain a "hall of fame" of top-performing candidates. 

Unlike DSPy's focus on prompt composition and application, HOPR emphasizes the development of automatic prompting methods by facilitating the implementation of concurrent strategies on the same problem. While being easily adaptible to new models, this allow a sain and reproducible comparative analysis of different prompt engineering approaches.

A key differentiator is HOPR's hybrid approach, which allows multiple optimization strategies to work in parallel, potentially discovering more effective prompts than single-strategy approaches. This makes it especially valuable for researchers studying prompt optimization methods and practitioners seeking to automatically optimize prompts for specific tasks. 

\subsection{Training pipeline}

\subsubsection{Dataset Organization}

The optimization process requires careful data partitioning to ensure robust evaluation and prevent overfitting. We divide each dataset into three distinct subsets:
\begin{itemize}[noitemsep, topsep=0pt]
    \item Training set: Used during prompt generation for strategy-specific optimization. APO leverages this set for error analysis and improvement, while the few-shot strategy uses it to select examples for in-context learning.
    \item Validation set: Employed during the optimization process to evaluate and compare generated prompts, enabling the selection of promising candidates for subsequent generations.
    \item Test set: Reserved exclusively for final evaluation, measuring generalization capability and tracking performance evolution across optimization steps.
\end{itemize}

\subsubsection{Population Management}

Each strategy is assigned a weight determining its contribution to the next generation's population. The number of candidates per strategy is calculated by multiplying these weights by the total population size. To maintain the exact desired population size, any remaining slots are allocated to the strategy with the highest weight. This weighted approach ensures:
\begin{itemize}[noitemsep, topsep=0pt]
    \item Balanced exploration across different optimization techniques
    \item Customizable strategy emphasis based on task requirements
    \item Consistent population size maintenance throughout generations
\end{itemize}

\subsubsection{Evaluation Process}

The evaluation of generated prompts follows a systematic approach:
\begin{itemize}[noitemsep, topsep=0pt]
    \item Initial evaluation on validation set to establish baseline performance.
    \item Generational evaluation to select promising candidates. Evaluations concerning advocated results in the paper where conducted twice, due to disparities in LLM response and their probabilistic answers construction.
    \item Final testing on the held-out test set to measure true generalization.
\end{itemize}
This structured pipeline ensures robust optimization while maintaining the flexibility to adapt to different tasks and requirements through adjustable strategy weights and evaluation parameters.

\subsubsection{Metrics}


\noindent\underline{\textbf{ETHOS multilabel dataset}}

For the multi-label classification task of the ETHOS dataset, we employ strict accuracy as our evaluation metric. A prediction is considered correct if and only if the set of predicted labels exactly matches the set of true labels, regardless of their order. Formally, for a sample with true labels $Y$ and predicted labels $\hat{Y}$, the binary accuracy is defined as:

\[
\text{accuracy}(Y, \hat{Y}) = \begin{cases} 
1 & \text{if } Y = \hat{Y} \\
0 & \text{otherwise}
\end{cases}
\]

where $Y$ and $\hat{Y}$ are treated as sets, meaning $\{a, b\} = \{b, a\}$. The final accuracy score is then computed as the average of these binary evaluations across all samples in the dataset. \\

\noindent\underline{\textbf{MMLU \& GPQA}}

For MMLU-Pro and GPQA datasets, we employ standard accuracy as our evaluation metric, where a prediction is considered correct if and only if it matches the correct answer. Formally, for a sample with true answer $y$ and predicted answer $\hat{y}$, the binary accuracy is defined as:

\[
\text{accuracy}(y, \hat{y}) = \begin{cases} 
1 & \text{if } y \equiv \hat{y} \\
0 & \text{otherwise}
\end{cases}
\]

where $\equiv$ denotes semantic equivalence rather than strict string matching. This equivalence consideration was necessary as these datasets provide multiple-choice answers in a standardized format (typically including punctuation marks like commas), but the LLM sometimes generated correct answers with slight formatting variations. To address this, we implemented an LLM-based evaluation system that validates semantic correctness, ensuring that superficial differences in formatting do not impact the accuracy assessment. The reliability of this approach was verified through manual inspection of a representative sample of model outputs.

\subsection{Experiments}

\indent \underline{\textbf{Datasets}}

300 samples were extracted to the orignal ETHOS dataset and separated in 3 subsets: 50 samples for the training set (used in the APO and the fewshot algorithms), 50 samples for the validation set (used for the selection of prompts at each generation) and 200 were used as test set to allow a meaningful comparison of different prompts while limiting the risks of overfitting on other subsets during the optimzation process. Those numbers were chosen as a tradeoff between the budget allowed to the optimization process and the typical size of the datasets which can be obtained in real life optimization tasks. Most results displayed in this paper use the ETHOS dataset. \\



\noindent\underline{\textbf{Models}}

We computed prompt optimization for several methods, which we reimplemented, respecting the original description made in their respective papers. In detail, APO \cite{pryzant2023apo}, OPRO\cite{yang2024opro} were used as baselines, along with a random mutator described above.

In the GAAPO implementation, prompt generation was distributed across strategies with the following proportions: random mutations accounted for 40\% of new prompts, while APO and OPRO each contributed 20\%. The remaining 20\% was equally divided between few-shot learning and crossover operations, each generating 10\% of new prompts.

These numbers were chosen as a trade-off between random prompt modifications (mutations and crossover), local prompt optimization (APO and OPRO) and in-context learning (fewshot). We deliberately choose to limit the importance of in-context learning as it has already been demonstrated that prompt efficiency scales with the number of given examples. Our goal here is to increase prompt efficiency for very small datasets to have a prompting method which can be used on real life prompts (comparison results are presented in section \ref{results:raw_scores}).

For each experiment, the number of generations and number of prompt generated at each generation was experimentally determined and will be justified in the Results section \ref{results:pop_size}. \\

\noindent\underline{\textbf{LLMs}}

The experimental setup employs two distinct Large Language Models (LLMs) for different aspects of the optimization process. 

For prompt generation, we utilize in most experiments \textit{DeepSeek-R1-distill-LLaMA-70B-versatile}, a state-of-the-art open-source LLM based on the LLaMA architecture. This model, accessed through Groq's inference platform, offers a balance between performance (with state-of-the-art performances on LLM tasks \cite{deepseekai2025deepseekr1}) and computational efficiency (with inference times sensibly lower using Groq platform \cite{groq2022inferencetime}). We compared the performance optimization obtained by this model to others in section \ref{results:llm_generator}.

For the target model to be optimized through our prompting process, we employ \textit{GPT-4o-mini} or \textit{llama3-8B-instant} \cite{grattafiori2024llama3herdmodels}. We decided to use 2 models to assess the difference in evolution performance (which can be seen in section \ref{results:llm_predictor}) across different experiment settings, arguing that a prompt optimization could be model dependent.

This configuration allows us to assess the generalizability of our prompt optimization approach while maintaining a clear separation between the prompt generation and evaluation phases of our methodology.\\

\noindent\underline{\textbf{Generalization}}

We evaluated our prompt optimization approach across several widely used datasets, with results presented in section \ref{results:datasets}. For each dataset, we maintained a consistent splitting strategy: 50 samples for training, 50 for validation, and up to 200 samples for testing (or the maximum available if fewer than 200 samples remained). This standardized approach, first validated on ETHOS, ensures fair comparison across different datasets while maintaining sufficient samples for reliable evaluation. Complementary datasets used for this study are presented in section \ref{section:data}. Note that for GPQA, only 98 samples were used in the testing set.\\

\noindent\underline{\textbf{Optimization of the selection process}}

To optimize the computational budget while maintaining effective prompt selection, we implemented and compared three different selection strategies (see section \ref{results:selection} for results): complete evaluation, successive halving, and bandit selection. These methods present different trade-offs between evaluation accuracy and computational efficiency.
For a representative scenario with a test dataset of 50 samples and a population of 50 prompts, the computational requirements vary significantly across methods. 
\begin{itemize}[noitemsep, topsep=0pt]
    \item Complete evaluation, which tests every prompt against every sample, requires 2,500 LLM calls ($50 \text{ prompts} \times 50 \text{ samples}$), providing exhaustive but computationally intensive evaluation.
    \item Successive halving \cite{schmucker2021sh} offers a more efficient approach by progressively eliminating underperforming prompts. In our implementation, we evaluate prompts on 20\% of the dataset at each iteration and eliminate 40\% of the lowest-performing prompts. This process continues until reaching a predetermined number of prompts. This strategy reduces the number of LLM calls to approximately 1,200, representing a 55\% reduction in computational cost compared to complete evaluation while maintaining robust selection pressure.
    \item The bandit selection method \cite{slivkins2024bandit} provides the most efficient tradeoff \cite{pryzant2023apo}, evaluating only 20 prompts on 15 samples over 5 iterations. This approach requires approximately 1,500 LLM calls ($20 \text{ prompts} \times 15 \text{ samples} \times 5 \text{ iterations}$), achieving a 40\% reduction in computational cost compared to complete evaluation. While this method samples less extensively, it leverages statistical efficiency to identify high-performing prompts.
\end{itemize}

These selection strategies offer different balances between evaluation thoroughness and computational efficiency, allowing practitioners to choose based on their specific constraints and requirements. Our empirical results suggest that both successive halving and bandit selection maintain effective prompt identification while significantly reducing computational overhead.

\section{Results \& Discussions} \label{section:results}
\subsection{Comparison with baselines} \label{results:raw_scores}

\begin{figure}[h]
    \centering
    \fbox{
    \includegraphics[width=0.9\textwidth]{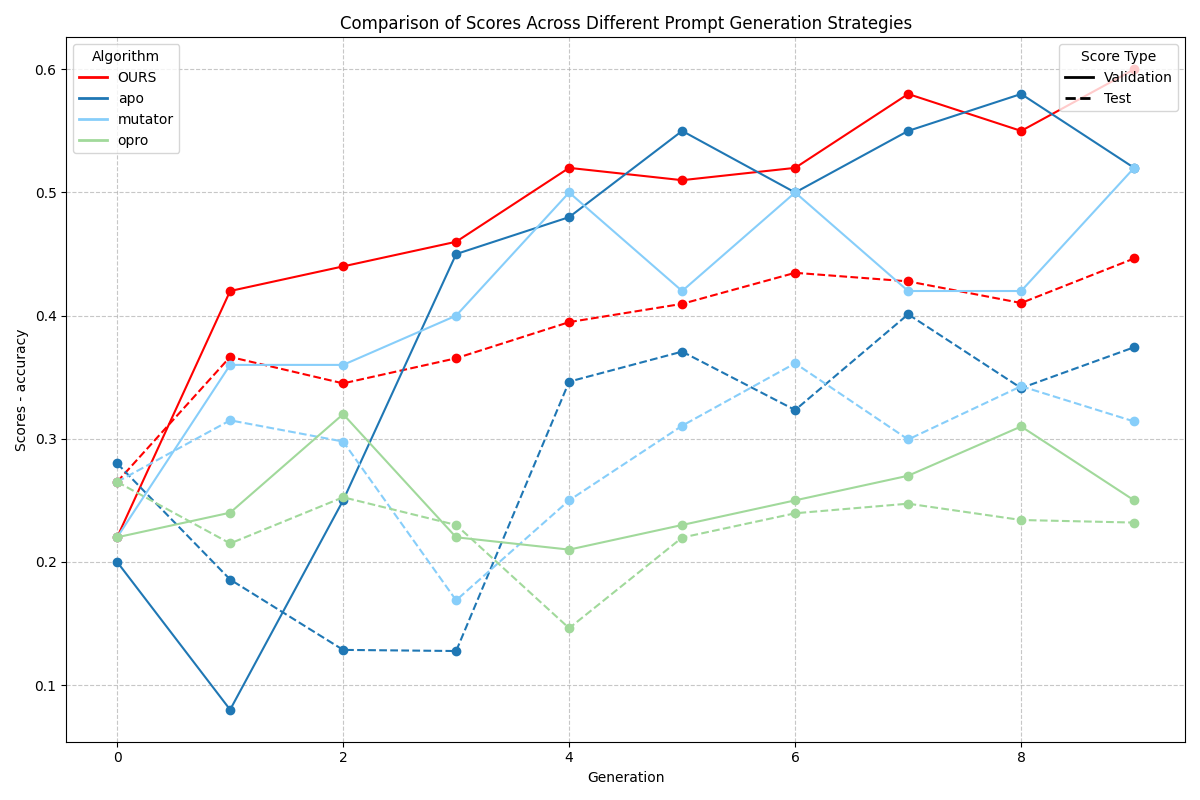}}
    \caption{Results obtained by using several prompt generation strategies. LLM-optimizer used: llama-3.1-8B}
    \label{fig:results}
\end{figure}

The experimental results demonstrate the effectiveness of our proposed GAAPO (Genetic Algorithm Assisted Prompt Optimization) approach on the ETHOS multilabel hate speech classification task. Figure \ref{fig:results} illustrates the validation performance across different prompt optimization strategies over multiple iterations, while Table \ref{tab:scores_llama_ethos} presents the final test and validation scores. Additionnally, obtained prompts are presented in table \ref{tab:obtained_prompts}.

\begin{table}[htbp]
\centering
\begin{tabular}{|p{2cm}|p{6cm}|p{6cm}|}
\hline
\textbf{Model} & \textbf{Validation score} & \textbf{Test score} \\
\hline
\textbf{\underline{Ours}} & \textbf{\underline{0.60}} & \textbf{\underline{0.46}} \\
\hline
APO & 0.52 & 0.38 \\
\hline
OPRO & 0.26 & 0.24 \\
\hline
Mutator & 0.52 & 0.34 \\
\hline
\end{tabular}
\caption{Test and validations scores for the ETHOS dataset. Comparison of performance of prompt optimization using llama3-8B-instant  \cite{grattafiori2024llama3herdmodels}}
\label{tab:scores_llama_ethos}
\end{table}

GAAPO demonstrates strong performance on the validation set, achieving a score of 0.46, which significantly surpasses baseline methods including OPRO (0.24), Mutator (0.34), and APO (0.38). The evolution curve in Figure \ref{fig:apo} shows GAAPO's ability to maintain consistent improvement throughout the optimization process.

A critical analysis of test and validation scores reveals an important phenomenon common to genetic algorithms: selection bias. This is particularly evident in APO's performance,  where results at some iterations highlight a very high difference score between test and validation sets (culminating at 0.3 for the 2nd generation). This extreme disparity illustrates how genetic algorithms can inadvertently optimize for specific test set characteristics rather than general problem-solving capabilities. GAAPO mitigates this selection bias through its diverse strategy portfolio, resulting in more balanced performance between test (0.60) and validation (0.46) scores, suggesting better generalization.

The lower performance of OPRO (test: 0.26, validation: 0.24) indicates that reinforcement learning-based approaches struggle with exploring vast prompt spaces effectively. The Mutator approach achieves intermediate results (test: 0.52, validation: 0.34), but still shows signs of selection bias with its significant test-validation gap. These observations highlight how selection bias can affect different optimization strategies to varying degrees, with GAAPO's hybrid approach providing the most robust defense against this common genetic algorithm limitation.

Moreover, we can see a difference in the original score of the models (at iteration 0, all scores should be identicals). However, due to the disparity in LLM performance and their probabilistic caracters, results are not always exactly consistent across time. 

\subsection{Model evaluation comparison} \label{results:llm_predictor}

\begin{figure}[h]
    \centering
    \fbox{
        \begin{minipage}{0.9\textwidth}
            \centering
            \includegraphics[width=\linewidth]{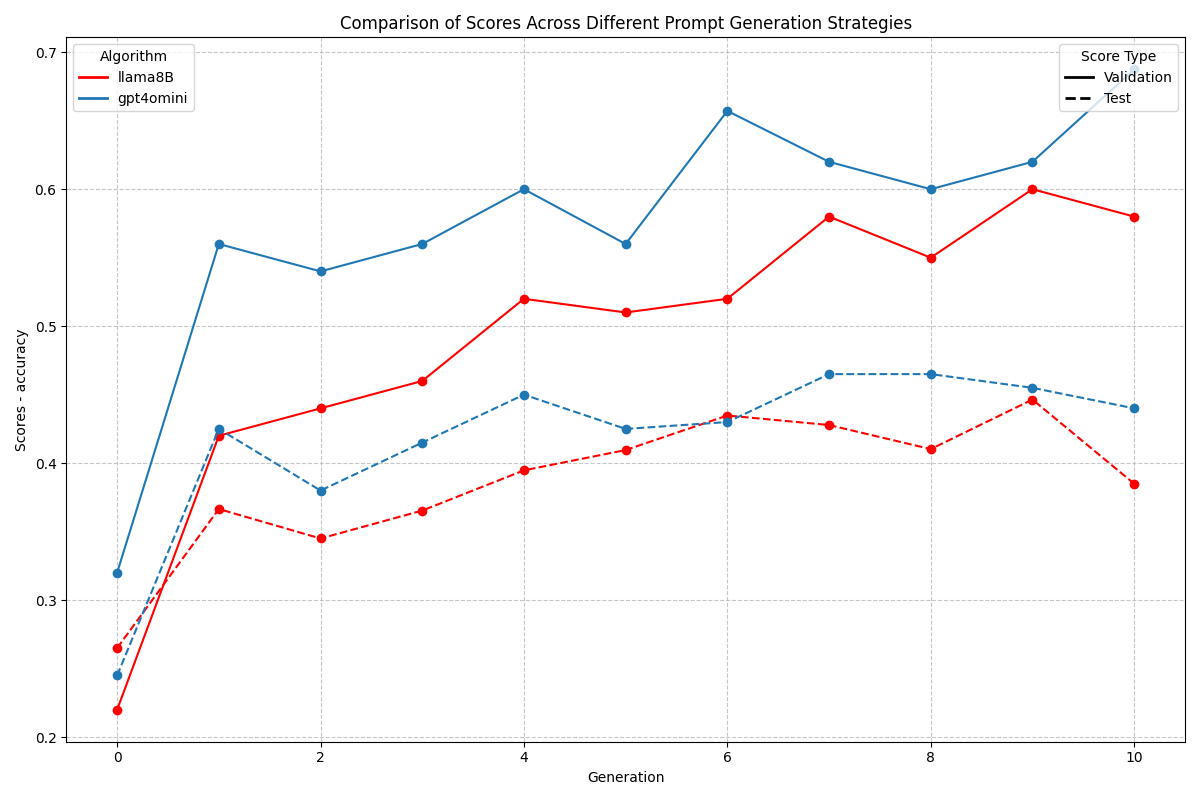}
        \end{minipage}
    }
    \caption{Comparison of optimization trajectories between GPT-4o-mini and LLaMA3-8B models on the ETHOS dataset for GAAPO. The plot shows the evolution of validation scores (solid lines) and test scores (dashed lines) across generations for both models.}
    \label{fig:comp_models}
    \vspace{-0.2cm}
\end{figure}

Comparing the optimization trajectories between GPT-4o-mini and LLaMA3-8B (displayed in figure \ref{fig:comp_models}) reveals few differences in how these models respond to prompt optimization. Both models show significant improvement from their initial performance, but their learning patterns and final achievements slightly differ.

Both GPT-4o-mini and LLaMA3-8B demonstrate stable optimization trajectories, with consistent learning patterns and similar generalization characteristics across generations. However, GPT-4o-mini achieves notably superior performance, reaching validation scores of up to 0.70 compared to LLaMA3-8B's 0.60. Both models maintain steady optimization paths with comparable stability in their generalization gaps.

Examining test scores reveals GPT-4o-mini's consistent edge in performance, maintaining a 0-0.05 point advantage over LLaMA3-8B throughout the optimization process. However, this superior performance must be interpreted with caution, as both models show signs of potential overfitting in later generations. The increasing gap between validation and test scores after generation 8 suggests that while GPT-4o-mini achieves better absolute performance, careful monitoring of generalization remains crucial for both models.

Given the higher performance metrics of GPT-4o-mini and comparable computational costs between the two models in our experimental setup, we selected GPT-4o-mini as our primary LLM-optimizer for subsequent experiments. This choice was driven by the quantitative advantages in optimization outcomes, while both models demonstrate equally reliable optimization stability.

\subsection{Influence of population size} \label{results:pop_size}

\begin{table}[htbp]
\centering
\begin{tabular}{|p{2cm}|p{3cm}|p{3cm}|p{3cm}|p{3cm}|}
\hline
\textbf{Population size} & \textbf{Number of generations} & \textbf{Test score} & \textbf{Validation score} & \textbf{Number of LLM calls} \\
\hline
 20 & 25 & 0.50 & 0.42 & 25000 \\
\hline
 30 & 17 & 0.56 & 0.50 & 25500\\
\hline
 40 & 13 & 0.62 & 0.46 & 24500\\
\hline
 50 & 10 & 0.68 & 0.46 & 25000\\
\hline
\end{tabular}
\caption{Test and validations scores for the ETHOS dataset. Comparison of different population size and number of generations for GPT-4o-mini. \cite{grattafiori2024llama3herdmodels}}
\label{tab:comp_pop_size}
\end{table}

We conducted experiments with varying population sizes while maintaining a comparable total number of LLM calls across configurations, as shown in Table \ref{tab:comp_pop_size}. The results demonstrate a clear trade-off between population size and the number of generations required. Larger populations (50 prompts) with fewer generations (10) achieve higher test scores (0.68) compared to smaller populations running for more generations (20 prompts, 25 generations, 0.50 test score).

While the configuration with 30 prompts shows the best validation score (0.50) and a smaller generalization gap, we opted for the 50-10 configuration for several practical advantages. First, larger populations enable better parallelization of prompt evaluation, significantly reducing wall-clock time. Second, this configuration aligns well with optimized selection strategy, which benefits from a larger pool of candidates to select from in each generation. 

However, the increased generalization gap in the 50-10 configuration (0.22 points between test and validation scores, compared to 0.08 points for 20-25) suggests a higher risk of overfitting. This observation indicates that while larger populations can explore the prompt space more effectively within fewer generations, they may require more robust validation strategies to ensure generalization. Despite this limitation, the practical benefits of faster convergence and improved parallelization potential make the 50-10 configuration our recommended choice for prompt optimization tasks.

\subsection{Prompt generators comparison} \label{results:generators}

We can now study in detail the prediction made by each prompt generator in GAAPO.

\begin{figure}[htbp]
    \centering
    \fbox{
        \begin{minipage}{0.9\textwidth}
            \centering
            \includegraphics[width=\linewidth]{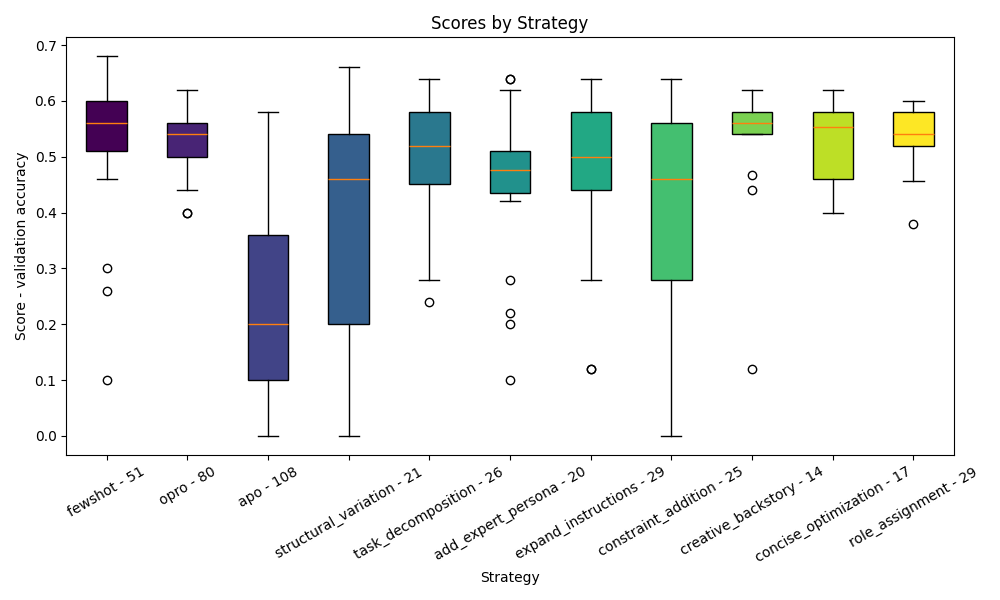}
        \end{minipage}
    }
    \caption{Performance distribution of individual prompt generation strategies  in GAAPO on the validation set. Model used: GPT-4o-mini.}
    \label{fig:strategies_boxplot}
    \vspace{-0.2cm}
\end{figure}

\begin{figure}[htbp]
    \centering
    \fbox{
        \begin{minipage}{0.9\textwidth}
            \centering
            \includegraphics[width=\linewidth]{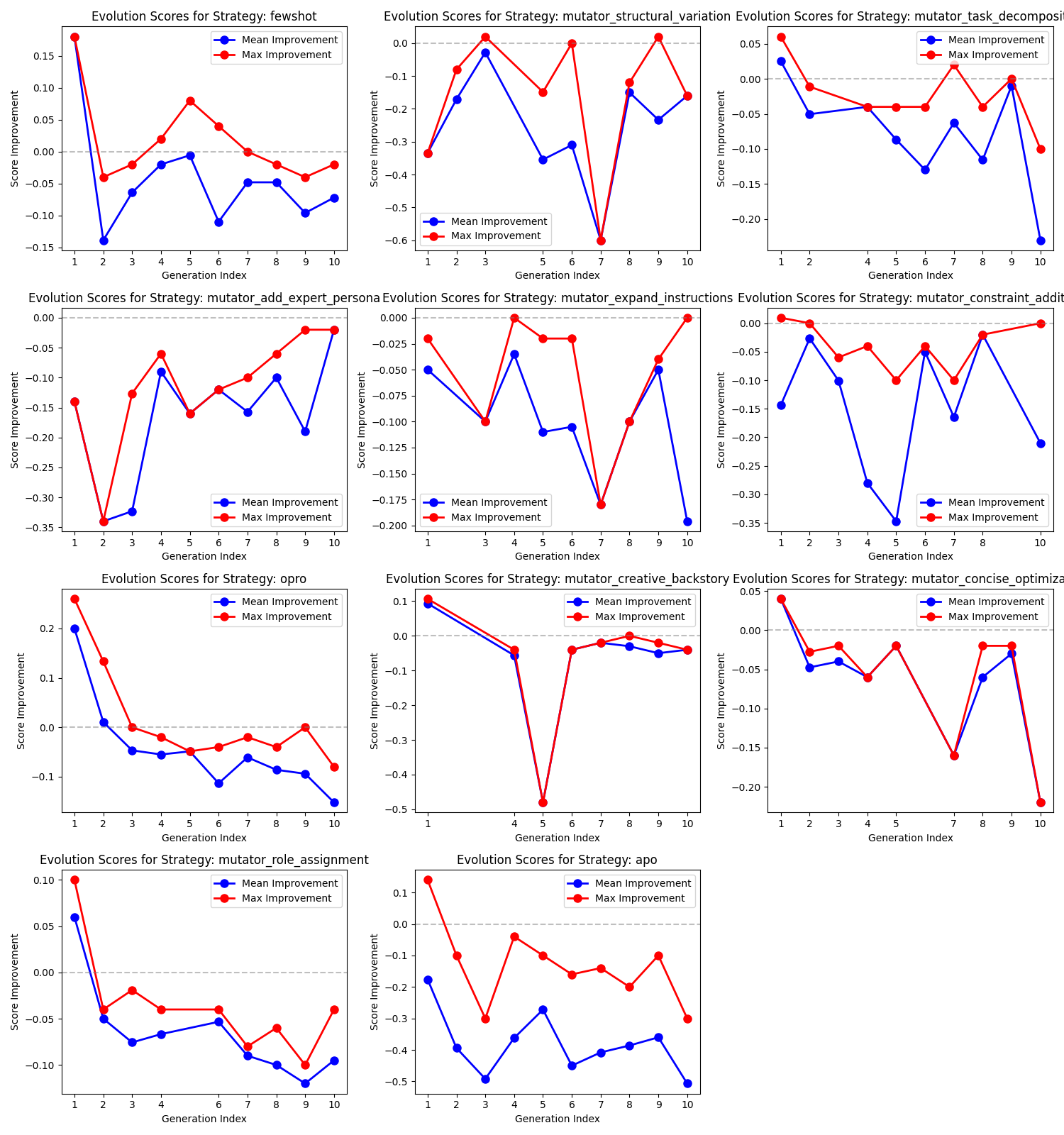}
        \end{minipage}
    }
    \caption{Evolution of improvement scores for each prompt generation strategy across generations. For each strategy, we track both mean improvement (blue) and maximum improvement (red) relative to parent prompts. Mean improvement represents the average score difference between generated prompts and their parents, while maximum improvement shows the best improvement achieved in each generation. Negative values indicate that generated prompts performed worse than their parents. Model used: GPT-4o-mini.}
    \label{fig:evol_boxplot}
    \vspace{-0.2cm}
\end{figure}

We conducted a detailed analysis of each prompt generator's performance in GAAPO through two complementary perspectives. Figure \ref{fig:strategies_boxplot} presents the overall distribution of validation scores for each strategy through boxplots, while Figure \ref{fig:evol_boxplot} tracks the improvement potential of each strategy across generations, showing both mean and maximum improvements in score relative to parent prompts.

To obtain these visualizations, we first aggregated all prompts generated by each strategy and analyzed their validation scores (Figure \ref{fig:strategies_boxplot}). Additionally, we computed the improvement in validation score between each generated prompt and its parent prompt across generations (Figure \ref{fig:evol_boxplot}), allowing us to understand not just absolute performance but also each strategy's ability to improve upon existing prompts.

The analysis reveals several key insights about strategy effectiveness and the importance of maintaining diversity in optimization approaches:
\begin{itemize}
    \item \textbf{\underline{Strategy Effectiveness and Stability:}} Few-shot learning demonstrates superior performance (median $\sim$0.57) with consistent results, as shown by its compact boxplot and positive improvement scores in early generations. This aligns with existing literature \cite{dong2024surveyICL}, highlighting the value of example-based learning. OPRO maintains strong and stable performance (median $\sim$0.55), though its evolution plot shows diminishing improvements over generations. \textit{role\_assignment} and \textit{concise\_optimization} show reliable performance with tight distributions, but their improvement potential decreases in later generations.

    \item \textbf{\underline{Evolution patterns:}} Most strategies show declining improvement potential over generations, with negative mean improvements in later stages, suggesting they work best in early exploration. APO's boxplot shows high variability (0.10-0.35), but its evolution plot reveals strong initial improvements followed by declining effectiveness, supporting its potential role as an early-stage optimizer. Few-shot learning uniquely maintains positive maximum improvements even in later generations, indicating sustained ability to generate beneficial variations.
    \item \textbf{\underline{Underperforming Strategies:}} Several mutation strategies, particularly \textit{structural\_variation} and \textit{task\_decomposition}, consistently show negative improvement scores across generations, suggesting limited effectiveness for the current task. However, completely removing these strategies could be counterproductive for two reasons:
    \begin{itemize}
        \item Task Dependency: Different tasks may benefit from different prompt modification approaches. What appears ineffective for one task might be crucial for another as every optimization task is learned in a different optimization space.
        \item Exploration Value: Even seemingly underperforming strategies contribute to maintaining genetic diversity, potentially enabling the discovery of novel promising prompt variations through combination with other approaches.
    \end{itemize}
    \item \textbf{\underline{Strategic Implications:}} The analysis suggests implementing a dynamic, task-adaptive strategy:
    \begin{itemize}
        \item Early generations: Leverage APO and mutation strategies for broad exploration
        \item Mid-generations: Emphasize few-shot learning and OPRO for stable improvements
        \item Later generations: Focus on strategies showing consistent positive improvements (few-shot,       \textit{role\_assignment}) for refinement
        \item Maintain a minimum weight for all strategies to preserve optimization flexibility across different tasks
    \end{itemize}
\end{itemize}

This comprehensive analysis reinforces the value of GAAPO's adaptable framework, which can accommodate varying strategy effectiveness across different tasks while maintaining the potential benefits of diverse optimization approaches. The framework's ability to dynamically adjust strategy weights while preserving all methods makes it particularly robust for general-purpose prompt optimization across diverse applications.

It should be notice that optimization methods tend to have descending curves which is logical: as we compare new prompts with their parent prompts, the task is more and more difficult (given that the reference prompt improves with the generations). Moreover, studies on other datasets tend to highlight the fact that different prompt optimization methods can perform very differently between tasks, highlightning the importance to keep methods in a general framework and the risk to select optimizers based on their results on a unique dataset.

\subsection{Model generators comparison} \label{results:llm_generator}

\begin{figure}[h]
    \centering
    \fbox{
        \begin{minipage}{0.9\textwidth}
            \centering
            \includegraphics[width=\linewidth]{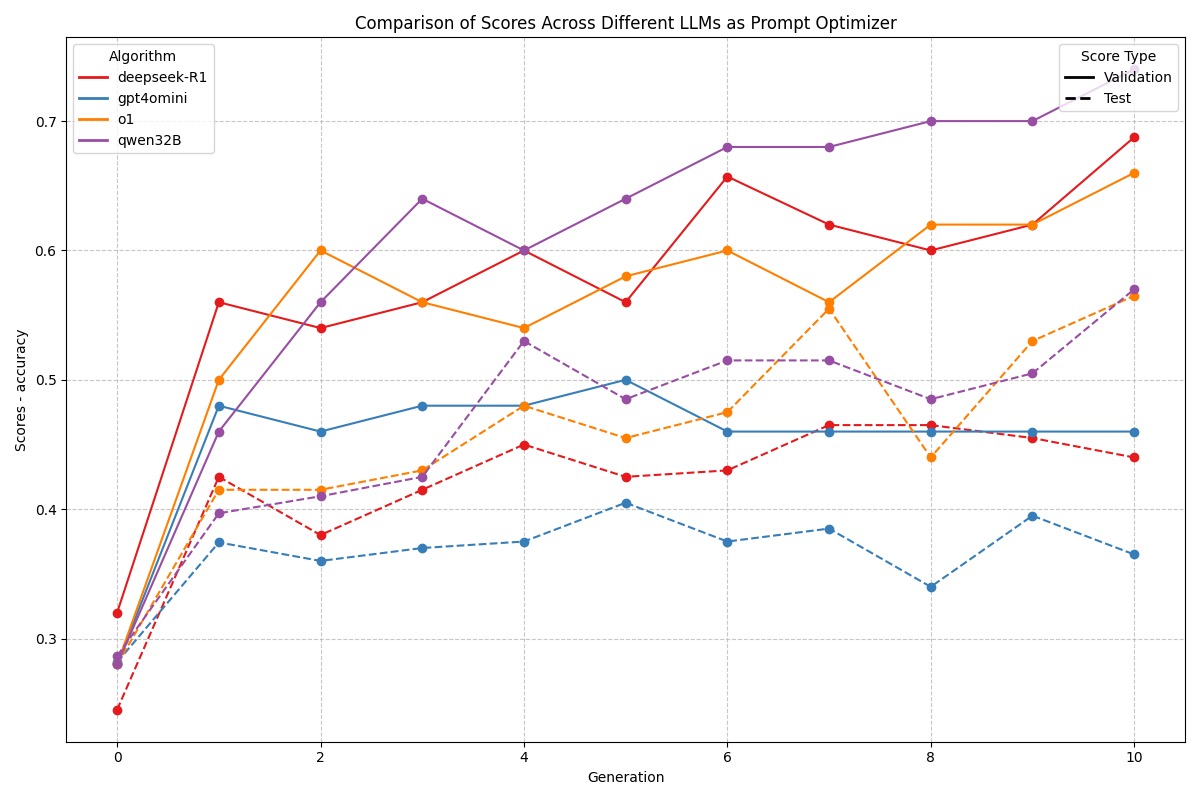}
        \end{minipage}
    }
    \caption{Comparison of different LLMs as prompt optimizers in GAAPO. The plot shows validation (solid lines) and test (dashed lines) scores across generations for four models: QwQ32B, DeepSeek-R1, O1, and GPT-4o-mini. While reasoning-specialized models achieve higher absolute scores, O1 demonstrates better generalization with smaller gaps between validation and test performance.}
    \label{fig:prompt_llm}
    \vspace{-0.2cm}
\end{figure}

The comparison of different language models as prompt optimizers reveals striking patterns (which can be seen in Figure \ref{fig:prompt_llm} in both performance and generalization capabilities. Most notably, reasoning-specialized models (QwQ32B and deepseek-R1) and O1 demonstrate superior performance compared to general-purpose models like GPT-4o-mini.

QwQ32B emerges as the top performer, showing consistent improvement in validation scores from an initial 0.28 to a remarkable 0.70 by generation 10. Its learning trajectory is particularly stable, with steady increases and minimal fluctuations. However, its test scores (dashed line) plateau around 0.55, indicating a significant generalization gap of approximately 0.15 points.

A particularly interesting comparison emerges between DeepSeek-R1 and O1 models. While both achieve strong final validation scores (0.68 and 0.65 respectively), O1 demonstrates notably better generalization characteristics. By generation 10, O1 maintains test scores around 0.55, nearly matching its validation performance, while DeepSeek-R1 shows a larger disparity with test scores around 0.45. This suggests that O1's optimization process, while slightly lower in absolute validation performance, produces more robust and generalizable prompts.

In contrast, GPT-4o-mini shows notably inferior performance. While it achieves quick initial improvement, its validation scores stagnate around 0.45-0.50 after generation 2, with minimal subsequent improvement. However, like O1, it maintains a smaller generalization gap between validation and test scores, suggesting more robust, if modest, optimization capabilities.

The evolution of scores across generations reveals an interesting pattern: while reasoning models continue to improve validation performance until the final generations, o1 maintains a more balanced improvement in both validation and test scores. This suggests that o1 might be particularly valuable for applications where generalization reliability is crucial, even if peak performance is slightly lower than specialized reasoning models.

These findings indicate that while reasoning-specialized models achieve higher absolute performance, o1 offers an attractive compromise between performance and generalization stability, potentially making it more suitable for practical applications where robust generalization is essential.

\subsection{Applications on other datasets} \label{results:datasets}

The experimental results across multiple datasets demonstrate both the effectiveness of our approach and the varying potential for prompt optimization across different tasks. Table \ref{tab:datasets_tested} presents validation scores for four distinct datasets, revealing several important patterns.

\begin{table}[htbp]
\centering
\begin{tabular}{|p{2cm}|p{2.5cm}|p{2.5cm}|p{2.5cm}|p{2.5cm}|}
\hline
\textbf{Dataset} & \textbf{ETHOS multilabel} & \textbf{MMLU-Pro engineering} & \textbf{MMLU-Pro Business} & \textbf{GPQA} \\
\hline
 Initialization & 0.28 & 0.39 & 0.72 & 0.38 \\
\hline
 APO & 0.44 & 0.45 & 0.73 & 0.42 \\
\hline
 OPRO & 0.38 & 0.44 & \textbf{\underline{0.76}} & \underline{0.43} \\
\hline
 Mutator & 0.40 & 0.43 & 0.735 & \underline{0.43} \\
\hline
 \textbf{OURS} & \textbf{\underline{0.46}} & \textbf{\underline{0.48}} & 0.74 & \underline{0.43} \\
\hline
\end{tabular}
\caption{Validations scores for different datasets. Models used: Deepseek-R1 as Prompt Generator and GPT-4o-mini as Optimizer.}
\label{tab:datasets_tested}
\end{table}

We can see on Table \ref{tab:datasets_tested} that our method achieves superior performance on datasets where prompt engineering shows significant potential for improvement. For the ETHOS multilabel classification task, we observe a substantial improvement from the initial score of 0.28 to 0.46, outperforming all baseline methods including APO (0.44), OPRO (0.38), and Mutator (0.40). Similarly, on the MMLU-Pro engineering dataset, our approach reaches 0.48, showing meaningful improvement over the initialization score of 0.39 and competing methods.

However, the results also reveal that not all tasks benefit equally from prompt optimization. The MMLU-Pro Business dataset, with its high initialization score of 0.72, shows minimal room for improvement, with our method and the Mutator achieving only marginal gains (0.73 and 0.735 respectively). This suggests that some tasks may already be well-aligned with LLMs' base capabilities, limiting the potential impact of prompt optimization. The GPQA dataset presents another interesting case where all optimization methods, including ours, achieve similar modest improvements (from 0.38 to 0.43), indicating that some tasks may have inherent complexity barriers that prompt optimization alone cannot overcome.

The varying effectiveness of prompt optimization across tasks can be attributed to multiple underlying factors. First, the overlap between an LLM's training data and the target dataset can create a ceiling effect - if similar examples were present in the training corpus, the model may already demonstrate near-optimal performance with simple prompts. Second, task-specific characteristics such as domain specificity and reasoning complexity influence the optimization potential; technical domains often benefit more from structured prompting than general knowledge tasks. Third, the nature of the required output (e.g., multiple-choice vs. multi-label classification) affects the scope for improvement through prompt engineering. Finally, the fundamental alignment between the task's requirements and the model's learned representations determines whether performance limitations can be addressed through prompt optimization alone or require more substantial interventions such as fine-tuning.

\subsection{Selection method comparison} \label{results:selection}

We conducted a comparative analysis of the three selection methods on the ETHOS dataset, evaluating their efficiency and performance trade-offs. The computational requirements varied significantly across methods: for a test set of 50 samples, the complete evaluation ("all") requires 2,500 LLM calls per generation, the bandit approach  approximately 1,500 calls, while successive halving (SH) uses only 1,500 calls per generation.

To ensure fair comparison, we also plotted results where the number of calls are equivalent between all methods. We adjusted the test size to 110 samples to obtain the right number of calls for both bandit and SH selection methods. 

\begin{figure}[h]
    \centering
    \fbox{
        \begin{minipage}{0.9\textwidth}
            \centering
            \includegraphics[width=\linewidth]{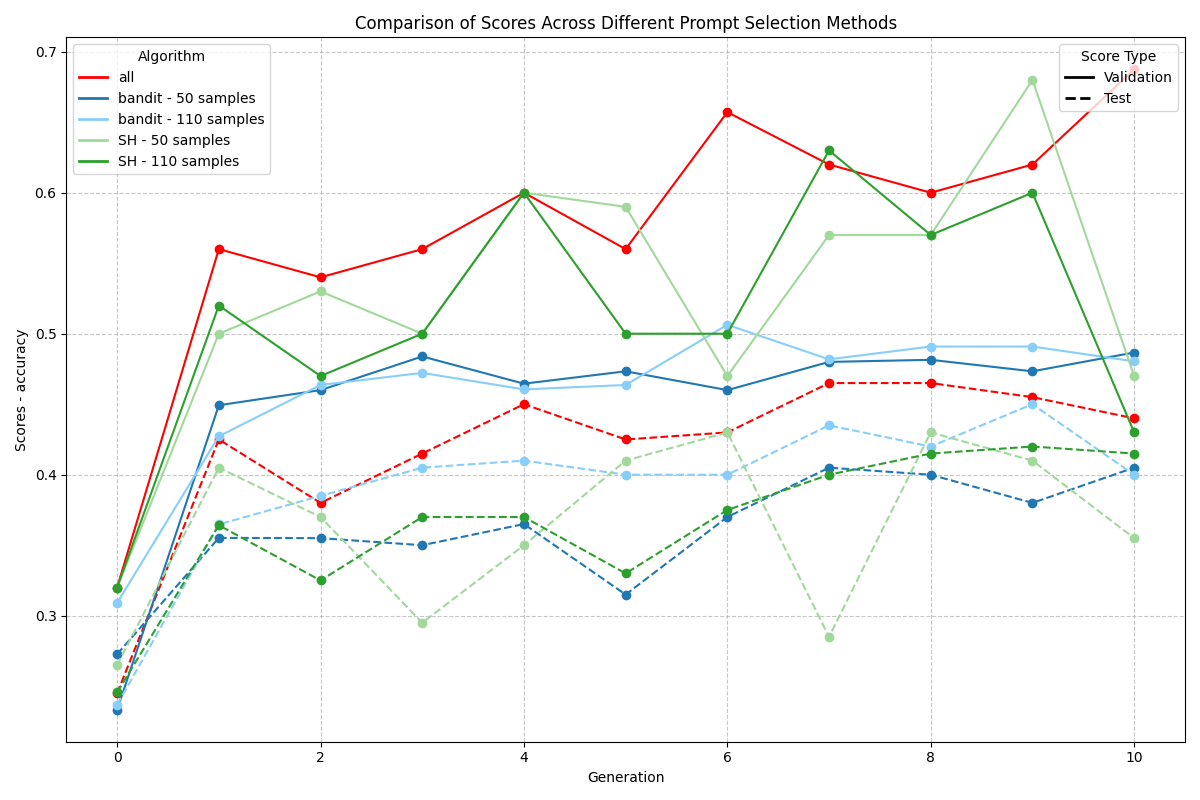}
        \end{minipage}
    }
    \caption{Comparison of different prompt selection strategies during GAAPO optimization. The plot shows the evolution of validation (solid lines) and test (dashed lines) scores across generations for different selection methods. Model used: GPT-4o-mini.}
    \label{fig:selection_comparison}
    \vspace{-0.2cm}
\end{figure}

Figure \ref{fig:selection_comparison} presents the evaluation for both validation and test scores for the 5 mentioned processes: "all", "bandit" with 50 samples, "bandit" with 110 samples, "SH" with 50 samples and "SH" with 110 samples.

The comparison of different selection strategies reveals compelling insights about the trade-offs between sample size, computational efficiency, and performance stability. The complete evaluation method ("all"), using 50 samples, achieves the highest validation scores (peaking at 0.68) but requires significantly more computational resources. However, our analysis demonstrates that increasing the sample size from 50 to 110 samples for alternative strategies does not necessarily lead to better performance, suggesting that efficient sampling is more crucial than sample size.

The bandit method emerges as particularly noteworthy, showing remarkable stability in both its 50 and 110 sample configurations. Despite using 40\% fewer LLM calls, it maintains consistent performance around 0.45-0.50 validation score with minimal fluctuations between generations. More importantly, the bandit approach exhibits a smaller generalization gap between test and validation scores, indicating better resistance to overfitting. However, we can observe a certain drop of performance between this selection method and "all".

In contrast, successive halving (SH) displays considerable volatility, especially evident in its performance spikes and drops across generations. While SH occasionally matches or exceeds the bandit's performance (reaching peaks around 0.60-0.70), its inconsistency makes it less reliable for practical applications. Interestingly, increasing the sample size for SH from 50 to 110 samples does not significantly mitigate this volatility, nor does it critically improve performances.

These findings suggest that while complete evaluation with 50 samples provides the highest absolute performance, the bandit approach with its reduced computational footprint and stable optimization trajectory offers an interesting alternative. The stability and efficiency of the bandit method, combined with its robust generalization characteristics, make it a choice for resource-conscious prompt optimization scenarios.

\section{Conclusion} \label{section:conclusion}
GAAPO (Genetic Algorithm Applied to Prompt Optimization) represents a significant advancement in prompt optimization, combining evolutionary strategies with established optimization methods. Our comprehensive evaluation demonstrates its effectiveness across multiple dimensions: superior validation performance with better generalization than baseline methods, efficient resource utilization through several prompt selection methods, and robust performance across different language models (GPT-4o-mini and LLaMA3-8B). The framework's modular architecture, incorporating multiple prompt generation strategies and selection methods, enables flexible adaptation to various tasks while maintaining optimization effectiveness.

However, several limitations warrant attention in future work. The framework shows increased generalization gaps with larger population sizes, suggesting the need for more sophisticated validation strategies. The computational overhead, while reduced through bandit selection, remains significant for resource-constrained applications. Future improvements could focus on developing more efficient prompt evaluation methods, incorporating active learning to reduce the number of required examples, and implementing adaptive population sizing strategies. Additionally, investigating the framework's effectiveness across a broader range of tasks and language models would enhance its generalizability and practical applicability.

\newpage

\section*{Acknowledgments}
We extend our sincere appreciation to Joël Belafa, Florence Armstrong, Lina Faik, Maxime Gobin and Charles-Albert Lehalle whose insightful comments and suggestions substantially enhanced this work. Their expertise and careful attention to detail helped us improve various aspects of the paper, from theoretical foundations to experimental validation. Their feedback not only improved the current work but also provided valuable directions for future research in this area.

\bibliographystyle{unsrt} 
\bibliography{bibliography}
\newpage

\section*{Annex}

\subsection*{MMLU-Pro dataset}\label{annex:mmlupro}

\begin{table}[h]
    \centering
    \renewcommand{\arraystretch}{1.3} 
    \begin{tabular}{|p{6cm}|p{6cm}|} 
        \hline
        \multicolumn{2}{|c|}{\textbf{MMLU-Pro: repartition of questions in the different categories}} \\ 
        \hline
        \textbf{Category} & \textbf{Number of questions} \\ 
        \hline
        Math & 1351 \\ 
        \hline
        Physics & 1299 \\ 
        \hline
        Chemistry & 1131 \\ 
        \hline
        Law & 1101 \\ 
        \hline
        Engineering & 969 \\ 
        \hline
        Economics & 844 \\ 
        \hline
        Health & 818 \\ 
        \hline
        Psychology & 798 \\ 
        \hline
        Business & 789 \\ 
        \hline
        Biology & 717 \\ 
        \hline
        Philosophy & 499 \\ 
        \hline
        Computer science & 410 \\ 
        \hline
        History & 381 \\ 
        \hline
        Others & 924 \\ 
        \hline
        Total & 12032 \\ 
        \hline
    \end{tabular}
    \caption{MMLUPro question repartitions.}
    \label{tab:mmlu}
\end{table}

\newpage

\subsection*{Prompt generation} \label{annex:prompt}

\begin{table}[h]
    \centering
    \begin{tabular}{|l|p{12cm}|}
        \hline
        \textbf{Mutation Type} & \textbf{Obtained Prompt} \\
        \hline
        Original prompt & 
        Answer this math problem: {{problem}} \\
        \hline 
        Expand Instructions & 
        Answer this math problem: {{problem}} 
        
        To solve this problem, follow these detailed step-by-step guidelines:
        \begin{itemize} [noitemsep, topsep=0pt]
            \item \textbf{Understand the Problem}: Identify what is being asked and determine the type of math problem (e.g., algebra, geometry, calculus).
            \item \textbf{Identify Known and Unknown Variables}: List all known values and the variables to solve.
            \item \textbf{Develop a Plan}: Choose the correct mathematical approach.
            \item \textbf{Execute the Plan}: Carry out calculations accurately, step by step.
            \item \textbf{Verify Your Solution}: Substitute the answer back into the problem to check correctness.
            \item \textbf{Reflect on the Solution}: Ensure clarity, correctness, and consider alternative solutions.
        \end{itemize}

        \textbf{Quality Criteria}: Clarity, correct notation, logical flow.  
        \textbf{Evaluation Metrics}: Accuracy, completeness, coherence.  
        \textbf{Common Mistakes}: Avoid calculation errors, logical jumps, and assumptions not supported by the problem. \\
        \hline
        Add Expert Persona & 
        You are a Fields Medalist specializing in complex mathematical problem-solving and theoretical mathematics.  
        With a Ph.D. in Mathematics and expertise in algebraic topology, differential geometry, and number theory,  
        you approach problems with analytical rigor and creative insight.  

        Answer this math problem: {{problem}} \\        
        \hline
        Task Decomposition & 
        To solve the given math problem, follow these steps:  
        \begin{itemize} [noitemsep, topsep=0pt]
            \item \textbf{Understand the Problem}: Carefully read the statement and identify relevant details.
            \item \textbf{Break Down the Problem}: Divide it into smaller, manageable parts.
            \item \textbf{Plan a Solution Strategy}: Determine which formulas or theorems apply.
            \item \textbf{Execute the Plan}: Perform calculations in a logical sequence.
            \item \textbf{Verify Intermediate Results}: Check calculations for accuracy at each step.
            \item \textbf{Combine Results}: Integrate partial solutions to derive the final answer.
            \item \textbf{Review and Validate}: Double-check correctness.
        \end{itemize}

        Answer the math problem: {{problem}} \\
        \hline
        Concise Optimization & 
        Solve: {{problem}} \\
        \hline
    \end{tabular}
    \caption{Mutation Types and Their Corresponding Prompts}
    \label{tab:mutation_prompts}
\end{table}

\begin{table}[h]
    \centering
    \begin{tabular}{|l|p{12cm}|}
        \hline
        \textbf{Mutation Type} & \textbf{Obtained Prompt} \\
        Structural Variation & 
        \textbf{Task Overview}: You are tasked with solving a mathematical problem using a structured approach.  

        \textbf{Problem Statement}:  
        \begin{itemize} [noitemsep, topsep=0pt]
            \item \textbf{Math Problem}: {{problem}} 
        \end{itemize}

        \textbf{Solution Strategy}:  
        \begin{itemize} [noitemsep, topsep=0pt]
            \item Understand the problem statement.
            \item Identify known variables and constraints.
            \item Determine required formulas or theorems.
            \item Solve systematically step-by-step.
            \item Verify your solution for accuracy.
        \end{itemize}

        \textbf{Common Mistakes to Avoid}:  
        \begin{itemize} [noitemsep, topsep=0pt]
            \item Misinterpreting the problem.
            \item Skipping verification steps.
            \item Incorrect formula application.
        \end{itemize}

        \textbf{Verification Steps}:  
        \begin{itemize} [noitemsep, topsep=0pt]
            \item Recheck each step for consistency.
            \item Ensure logical correctness.
        \end{itemize}

        \textbf{Output Format} (JSON):  
        \begin{verbatim}
        {
            "solution": "your detailed solution steps",
            "final_answer": "your final answer"
        }
        \end{verbatim} \\
        \hline
        Creative Backstory & 
        In the year 2147, aboard the interstellar vessel \textbf{Math Explorer},  
        you are the chief mathematician responsible for ensuring safe passage.  
        The ship's navigation system has encountered a critical error.  

        Only by solving the following math problem can you recalibrate the system  
        and prevent a catastrophic collision with a rogue asteroid.  
        The fate of the crew and the success of the mission depend on your expertise.  

        Answer this math problem: {{problem}}\\
        \hline
        Constraint Addition & 
        Answer this math problem:{{problem}}  
        \begin{itemize} [noitemsep, topsep=0pt]
            \item Do not use any numbers greater than 10.
            \item Explain your solution as if teaching a 10-year-old using biological analogies.
            \item Solve within 5 minutes.
            \item Do not include algebraic expressions or terminology.
            \item Present the final answer in haiku format.
        \end{itemize} \\
        \hline
        Role Assignment & 
        You are a team of mathematicians working collaboratively to solve the problem.  

        \begin{itemize} [noitemsep, topsep=0pt]
            \item \textbf{Critic}: Analyze the problem’s complexity and identify possible challenges: {{problem}}.
            \item \textbf{Problem Solver}: Devise a step-by-step strategy.
            \item \textbf{Teacher}: Explain the solution clearly for a novice audience.
        \end{itemize}

        Your response should integrate insights from each role, providing a thorough yet accessible solution. \\
        \hline
    \end{tabular}
    \caption{Mutation Types and Their Corresponding Prompts (end)}
    \label{tab:mutation_prompts}
\end{table}

\clearpage
\newpage

\end{document}